\documentclass[conference]{IEEEtran}
\IEEEoverridecommandlockouts

\usepackage{amsmath,amssymb,amsthm}
\usepackage{graphicx}
\usepackage{hyperref}
\usepackage{listings}
\usepackage{algorithm}
\usepackage{algpseudocode}
\usepackage{multirow}
\usepackage{booktabs}
\usepackage{caption,subcaption}
\usepackage{microtype}
\usepackage{xcolor}
\usepackage{array}
\usepackage{float}
\usepackage{enumitem}
\usepackage{authblk}
\usepackage{lipsum}

\hypersetup{
    colorlinks=true,
    allcolors=blue,
    pdftitle={Interpretability-Guided Bi-objective Optimization},
    pdfauthor={Kasra Fouladi, Hamta Rahmani},
}

\newcommand{\R}{\mathbb{R}}
\newcommand{\E}{\mathbb{E}} 
\newcommand{\Loss}{\mathcal{L}}
\newcommand{\HLoss}{\mathcal{H}}
\newcommand{\Var}{\mathrm{Var}}

\newtheorem{remark}{Remark}

\title{Interpretability-Guided Bi-objective Optimization: \\ Aligning Accuracy and Explainability}

\author[1]{Kasra Fouladi\thanks{Corresponding author: \texttt{k4sr405@gmail.com}. B.Sc. student, Department of Computer Science, Iran University of Science and Technology.}}
\author[1]{Hamta Rahmani\thanks{\texttt{hamtar693@gmail.com}. B.Sc. student, Department of Computer Science, Iran University of Science and Technology.}}
\affil[1]{Department of Computer Science, Iran University of Science and Technology, Tehran, Iran}
\affil{\textbf{GitHub}: \url{https://github.com/hamtarahmani/IGBO}}

\begin{document}

\maketitle

\begin{center}
\small\textit{Note: AI writing assistance was utilized for improved readability while preserving all technical integrity.}
\end{center}

\begin{abstract}
This paper introduces Interpretability-Guided Bi-objective Optimization (IGBO), a framework that trains interpretable models by incorporating structured domain knowledge via a bi-objective formulation. IGBO encodes feature importance hierarchies as a Directed Acyclic Graph (DAG) via Central Limit Theorem-based construction and uses Temporal Integrated Gradients (TIG) to measure feature importance. The framework employs a \emph{Relative Importance Score} \(H_k(\mathbf{X},\theta)\) that quantifies the normalized cumulative attribution of each feature over time. We propose a geometric projection mapping $\mathcal{P}$ for combining task and interpretability gradients, and prove convergence to Pareto-stationary points. To address the Out-of-Distribution problem in TIG computation, we outline an Optimal Path Oracle architecture, which we leave for future work. Central Limit Theorem-based construction of the interpretability DAG provides statistical guarantees on acyclicity and transitivity, with an unconditional guarantee for the median threshold and conditional guarantees for higher confidence levels.

\vspace{0.5em}
\noindent\textbf{Keywords:} Interpretable Machine Learning, Bi-objective Optimization, Explainable AI (XAI), Temporal Integrated Gradients, Optimal Path Oracle, Central Limit Theorem
\end{abstract}

\section{Introduction}
Deep learning models have achieved remarkable success in sequential prediction tasks across domains such as healthcare and finance. However, their inherent complexity often renders them opaque ``black boxes,'' limiting trust and deployment in safety-critical scenarios where understanding the model's reasoning is as important as its accuracy. Post-hoc explanation methods, including Integrated Gradients \cite{sundararajan2017axiomatic} and LIME \cite{ribeiro2016should}, attempt to illuminate this reasoning by analyzing trained models. A fundamental limitation persists: these methods are applied \emph{after} training, providing no guarantee that the model's internal decision-making aligns with established domain knowledge or desired behavioral constraints during the learning process itself \cite{mironicolau2024comprehensive}. Consequently, a model may achieve high accuracy by exploiting spurious, non-causal, or ethically problematic correlations in the data—a risk unacceptable in high-stakes applications.

To build more trustworthy systems, interpretability must be integrated directly into the training objective. A common approach adds a regularization term penalizing undesirable behavior. However, such methods often treat interpretability as a monolithic, soft constraint, offering practitioners limited control over the precise trade-off between accuracy and interpretability. More critically, they lack a formal mechanism to enforce \emph{structured}, \emph{relational} constraints that reflect complex domain expertise, such as hierarchies of feature importance. Additionally, gradient-based attribution methods like Integrated Gradients suffer from the OOD problem when computing gradients along straight-line paths between inputs and baselines.

This work proposes a principled framework to address these gaps: \textbf{Interpretability-Guided Bi-objective Optimization (IGBO)}. IGBO formalizes interpretable model training as a bi-objective optimization problem where the main model $F_\theta$ is trained to simultaneously minimize a primary task loss $\Loss(\theta)$ and an interpretability loss $\HLoss(\theta)$ via a geometric projection mapping $\mathcal{P}$ (Algorithm \ref{alg:projected_update_complete}). The interpretability loss is derived from two key components: (1) a \textbf{Temporal Integrated Gradients (TIG)} metric that provides a differentiable measure of feature importance over time, and (2) a \textbf{Directed Acyclic Graph (DAG)} defined via Central Limit Theorem-based construction (Section \ref{sec:graph_clt}) that encodes permissible relative importance relationships between features (e.g., feature $A$ should be more important than feature $B$). The feature importance is captured by a \textbf{Relative Importance Score} $H_k(\mathbf{X},\theta)$, a signed, normalized measure of each feature's contribution. We also discuss an \textbf{Optimal Path Oracle} to improve TIG robustness, which we leave as a direction for future work.

\textbf{Gradient-Based Optimization:} The bi-objective problem requires navigating trade-offs between competing gradients. We employ a geometric gradient combination rule based on the cosine similarity between gradient vectors. When gradients are aligned, we use a convex combination; when they conflict, we combine their orthogonal components. This approach provides descent guarantees for both objectives while avoiding the need for secondary optimization subroutines.

\textbf{Application:} The IGBO framework not only provides immediate interpretability guarantees but also opens avenues for human-in-the-loop model alignment and task-specific model specialization, which we explore in later sections (Section \ref{sec:methodology}, \ref{sec:discussion}).

\textbf{Contributions:} This paper makes the following contributions:
\begin{itemize}[noitemsep, topsep=2pt]
    \item We introduce the \textbf{IGBO framework}, which integrates structured domain knowledge (via a DAG) into gradient-based learning through a differentiable TIG-based \emph{Relative Importance Score} $H_k(\mathbf{X}, \theta)$, addressing both interpretability constraints and providing a principled bi-objective optimization method.
    \item We propose a \textbf{geometric projection mapping} $\mathcal{P}$ that gives a closed-form solution for the two-objective steepest descent direction, and we prove convergence to Pareto-stationary points (Theorems \ref{thm:aligned_complete}--\ref{thm:convergence_pareto}) and robustness to gradient noise (Theorem \ref{thm:noise_characterization}).
    \item We establish \textbf{statistical guarantees} for the CLT-based DAG construction, including unconditional acyclicity for the median threshold and conditional acyclicity/transitivity for higher confidence levels (Section \ref{sec:graph_clt}).
\end{itemize}

\section{Related Work}
Our work synthesizes concepts from interpretable AI, gradient-based multi-objective optimization, robust attribution methods, and the geometry of loss landscapes. We position IGBO within these interconnected research streams.

\textbf{Integrating Interpretability into Training:} Moving beyond post-hoc explanation methods like LIME \cite{ribeiro2016should} and SHAP \cite{lundberg2017unified}, recent approaches integrate interpretability directly into the training loop. Methods like Right for the Right Reasons \cite{ross2017right} regularize input gradients to align with human annotations, focusing on local, per-instance explanations. Concept Bottleneck Models (CBMs) \cite{koh2020concept} enforce interpretability by routing predictions through human-specified concepts, requiring concept-level labels. Other works use symbolic constraints or logic rules to guide learning \cite{marra2019lyra}. In contrast, IGBO enforces \emph{global, relational constraints} on aggregate feature importance via a domain-expert DAG, without requiring per-instance annotations or explicit concept labels. This provides a structured prior over model behavior, complementing approaches that integrate domain knowledge to ensure explanation fidelity \cite{mironicolau2024comprehensive,rudin2019stop}. Our work also relates to methods imposing sparsity or simplicity constraints but differs in its focus on explicit, relational importance hierarchies.

\textbf{Multi-Objective Gradient-Based Optimization:} IGBO relies on gradient-based techniques for multi-objective optimization. The Multi-Gradient Descent Algorithm (MGDA) \cite{desideri2012multiple, fliege2019steepest} provides a general framework for finding descent directions common to multiple objectives, often requiring solving a quadratic program. For the prevalent two-objective case, IGBO's geometric update rule provides a closed-form solution equivalent to MGDA, eliminating the optimization subroutine. This connects IGBO to gradient-based multi-task learning \cite{sener2018multi}, where conflicting gradients are a central challenge, and to Pareto-frontier tracing methods \cite{lin2020pareto}. Recent work on multi-objective optimization includes adaptive weighting schemes \cite{navon2022multi} and gradient manipulation techniques \cite{liu2021conflict}, but these typically do not consider interpretability-specific constraints as structured objectives.

\textbf{Gradient-based Attribution and Robust Path Methods:} Our TIG builds directly upon Integrated Gradients (IG) \cite{sundararajan2017axiomatic}, inheriting its axiomatic properties. Recent work has identified limitations in IG, particularly its sensitivity to the choice of baseline \cite{sturmfels2020visualizing} and the fact that straight-line integration paths often traverse OOD regions, leading to unreliable attributions \cite{hess2020middle}. Several approaches attempt to address these issues: Expected Gradients \cite{erion2019learning} averages over multiple baselines; Blurring Integrated Gradients \cite{xu2020explainable} applies smoothing; and adversarial training methods aim to produce more robust attributions \cite{giove2023robust}. We leave the training of an Optimal Path Oracle to future work, which could learn a data-manifold-aware integration path, fundamentally changing how the attribution integral is computed. This approach is conceptually related to works exploring shortest paths on manifolds for explanations \cite{garreau2020explaining} but would differ in its fully differentiable, learned formulation and integration into a bi-objective training loop.

\textbf{Geometry of Loss Landscapes and Gradient Alignment:} Understanding and leveraging the geometry of high-dimensional loss landscapes is crucial for optimization \cite{kiselev2024unraveling}. The geometric update rule in IGBO explicitly operates within the low-dimensional subspace spanned by two gradients. This connects to research on gradient conflict analysis in multi-task learning \cite{yu2020gradient}, where gradient directions are manipulated to improve convergence. The use of orthogonal gradient components when objectives conflict relates to techniques for escaping saddle points \cite{dauphin2014identifying} and methods that project gradients to find mutually beneficial directions \cite{schafer2021continuous}. IGBO's approach is tailored specifically for navigating the accuracy-interpretability trade-off with clear geometric intuition, providing a practical instance of these geometric principles.

\textbf{Probabilistic Graph Construction:} Our method for constructing the interpretability DAG via Central Limit Theorem approximation (Section \ref{sec:graph_clt}) relates to statistical methods for causal discovery \cite{spirtes2000causation} and confidence-based edge orientation \cite{kalisch2007estimating}. However, unlike causal discovery methods that infer structure from data, our approach uses statistical confidence to operationalize expert knowledge about feature importance hierarchies, with theoretical guarantees on acyclicity (Corollary \ref{cor:acyclic}) and transitivity (Theorem \ref{thm:transitivity}) under appropriate conditions.


\section{Methodology}
\label{sec:methodology}

The IGBO framework formalizes interpretable model training as a bi-objective optimization problem. In this work we focus on training the main sequential model $F_\theta$ to balance \emph{predictive accuracy} and \emph{interpretability constraints}; the training of an Optimal Path Oracle (Section \ref{subsec:routing_oracle}) is left for future investigation. This section formalizes the interpretability prior, introduces the TIG-based importance measure, and details the proposed Routing Oracle as a future extension.

\subsection{Interpretability Prior Definition}
\label{sec:prior}

\paragraph{Sequential Model Definition} Let $\mathcal{X} = \R^{T \times d}$ denote the input space of multivariate time series of length $T$ and feature dimension $d$. A sequential model processes an input $\mathbf{X} = [\mathbf{x}_1, \dots, \mathbf{x}_T] \in \mathcal{X}$ through a parameterized transition function $g_\theta$ and an output function $f_\theta$. The hidden state $\mathbf{h}_t \in \R^h$ at time $t$ evolves as:
\begin{equation}
\mathbf{h}_{t} = g_\theta(\mathbf{x}_t, \mathbf{h}_{t-1}), \quad t = 1,\dots,T,
\end{equation}
with $\mathbf{h}_0 = \mathbf{0}$. The output at time $t$ is given by $y_t = f_\theta(\mathbf{x}_t, \mathbf{h}_{t-1}) \in \mathbb{R}$. Consequently, the full sequence-to-sequence model $F_\theta: \mathcal{X} \to \mathbb{R}^T$ maps an input sequence to an output sequence, where its $t$-th component is $[F_\theta(\mathbf{X})]_t = f_\theta(\mathbf{x}_t, \mathbf{h}_{t-1})$.

\paragraph{Temporal Integrated Gradients.} 
To obtain a differentiable measure of feature importance over time, we extend Integrated Gradients~\cite{sundararajan2017axiomatic} to sequential data. For a given input $\mathbf{X}$ and a baseline $\mathbf{X}'$, the importance of feature $k$ at time $t$ along a path $\gamma: [0,1] \to \mathcal{A}$ connecting $\gamma(0)=\mathbf{X}'$ to $\gamma(1)=\mathbf{X}$ is defined as:

\begin{equation}
\label{eq:tig_def_general}
\text{TIG}_{t,k}^{\gamma}(\mathbf{X}, \theta) = \int_{0}^{1} 
\frac{\partial F_\theta(\gamma(\alpha))}{\partial \mathbf{X}_{t,k}} \cdot 
\frac{d\gamma_{t,k}(\alpha)}{d\alpha} \, d\alpha .
\end{equation}

For the special case of a straight-line path $\gamma_{\text{lin}}(\alpha) = \alpha\mathbf{X} + (1-\alpha)\mathbf{X}'$, this simplifies to the standard IG formulation:

\begin{equation}
\label{eq:tig_def_linear}
\text{TIG}_{t,k}^{\gamma_{\text{lin}}}(\mathbf{X}, \theta) = (\mathbf{X}_{t,k} - \mathbf{X}'_{t,k}) \int_{0}^{1} 
\frac{\partial F_\theta(\gamma_{\text{lin}}(\alpha))}{\partial \mathbf{X}_{t,k}} \, d\alpha .
\end{equation}

This integral aggregates the sensitivity of the model's output to feature $(t,k)$ along the entire path $\gamma$. The standard implementation uses the straight-line path $\gamma_{\text{lin}}(\alpha) = \alpha\mathbf{X} + (1-\alpha)\mathbf{X}'$. However, a major drawback of this linear interpolation is that intermediate points $\gamma_{\text{lin}}(\alpha)$ often lie outside the training data distribution (OOD). Since $F_\theta$ is not trained to behave consistently in OOD regions, the resulting gradients—and hence the attributions—can be unreliable and noisy.

\subsubsection{Routing Oracle: Learning Data-Manifold-Aware Integration Paths}
\label{subsec:routing_oracle}

\paragraph{Motivation and Design Philosophy} 
To circumvent the OOD problem inherent in straight-line integration paths, we propose a \emph{Routing Oracle} (also referred to as a \emph{Pathfinder}) that learns to generate piecewise-linear paths adhering to the data manifold while minimizing total path length. By ensuring all intermediate integration points reside within regions where the model has been properly trained, the oracle yields stable and reliable gradient computations for TIG. In this paper we describe the oracle architecture and training procedure but leave its full implementation and integration into the training loop as future work; all experiments use the simple linear path.

\paragraph{Flexible Implementation Options} 
The oracle is designed with practical deployment in mind, offering several configuration choices:
\begin{itemize}
    \item \textbf{Baseline selection:} The baseline $\mathbf{X}'$ can be chosen per sample (e.g., using dataset statistics) or set globally (e.g., a zero sequence). For maximum simplicity, $\mathbf{X}'$ can be fixed as a hyperparameter.
    \item \textbf{Path complexity control:} The number of anchor points $K$ can be tuned according to the available computational budget. Larger $K$ allows the path to better follow the data manifold but increases inference cost. For deterministic high-throughput deployment, $K$ can also be fixed as a hyperparameter.
\end{itemize}
This flexibility lets practitioners balance attribution quality against computational cost for their specific application.

\paragraph{Oracle Architecture and Recurrence Structure} 
The routing oracle is implemented as a recurrent neural network $G: \mathcal{X} \times \mathcal{X} \times \mathbb{N} \to \mathcal{X}^*$ that, given a baseline $\mathbf{X}'$, a target input $\mathbf{X}$, and a desired number of steps $K$, generates a sequence of $K$ anchor points:

\begin{align*}
[\mathbf{p}_1, \dots, \mathbf{p}_K] = G(\mathbf{X}', \mathbf{X}, K; \theta_G).
\end{align*}

The architecture employs a recurrence that explicitly accounts for the remaining path length. Let $\mathbf{h}_i \in \R^h$ be a hidden state. Starting with $\mathbf{h}_0 = \mathbf{0}$ and $\mathbf{p}_0 = \mathbf{X}'$, the $i$-th point is produced as:

\begin{align*}
&\mathbf{p}_i = r^\text{output}_{\theta_G}\bigl(K-i+1,\; \mathbf{X},\; \mathbf{p}_{i-1},\; \mathbf{h}_{i-1}\bigr),\\
&\mathbf{h}_i = r^\text{hidden}_{\theta_G}\bigl(K-i+1,\; \mathbf{X},\; \mathbf{p}_{i-1},\; \mathbf{h}_{i-1}\bigr),
\end{align*}

where $r$ is a learned function. The explicit inclusion of the remaining-step count $(K-i+1)$ allows the network to plan the path efficiently and adapt its behavior as the endpoint approaches.

\paragraph{Two-Phase Training Procedure} 
Training is performed in two distinct phases to ensure stability and reuse existing knowledge of the data distribution:
\begin{enumerate}
    \item \textbf{Discriminator pre-training:} We employ the discriminator $D(\cdot; \theta_D)$ from a pre-trained Generative Adversarial Network (GAN) as a frozen \emph{validity assessor}. This discriminator, already trained to distinguish real data from synthetic samples, provides a reliable estimate of $\Pr(\mathbf{p} \in \mathcal{A})$ without requiring additional training.
    \item \textbf{Oracle training:} With $D$ fixed, we train $G$ to minimize a composite loss that combines path shortness and point validity:
    \begin{equation}
    \mathcal{L}_G(\theta_G) = \mathbb{E}_{(K, \mathbf{X}',\mathbf{X})} \bigl[ \mathcal{L}_{\text{path}}, \, \mathcal{L}_{\text{valid}} \bigr],
    \end{equation}
    where
    \begin{align}
    \begin{split}
    &\mathcal{L}_{\text{path}} = \\ &\frac{\|\mathbf{X}' - \mathbf{p}_1\|^2 + \|\mathbf{X} - \mathbf{p}_K\|^2 + \sum_{i=1}^{K-1} \|\mathbf{p}_{i+1} - \mathbf{p}_i\|^2}{K + 1},
    \end{split} \\
    &\mathcal{L}_{\text{valid}} = \frac{-1}{K}\sum_{i=1}^{K} \log D(\mathbf{p}_i; \theta_D).
    \end{align}
    Training the oracle is a bi-objective optimization problem, which can be approached with the same geometric update rule used for the main model.
\end{enumerate}

\paragraph{Integration Strategy}
For the actual TIG computation with oracle-generated paths, we approximate the integral in Eq.~\eqref{eq:tig_def_general} using numerical integration. Given $M$ integration points $\mathbf{q}_1,\dots,\mathbf{q}_M$ along the path (including oracle anchors and interpolated points), we compute:

\begin{equation}
\label{eq:tig_oracle_final_correct}
\text{TIG}_{t,k}^{\text{Oracle}}(\mathbf{X}, \theta) \approx 
\sum_{j=1}^{M-1} \frac{\partial F_\theta(\mathbf{q}_j)}{\partial \mathbf{q}_{j,t,k}} \cdot 
(\mathbf{q}_{j+1,t,k} - \mathbf{q}_{j,t,k}),
\end{equation}

which corresponds to a Riemann sum approximation of the line integral along the piecewise-linear path connecting the oracle's anchor points.

\paragraph{Transferability and Reusability} 
A significant practical advantage of the Routing Oracle is its transferability across models. Once trained on a representative dataset $\mathcal{D}$, the same oracle can be applied to compute TIG for \emph{any} model $F_\theta$ trained on $\mathcal{D}$, without retraining. This enables:
\begin{itemize}
    \item \textbf{Consistent evaluation:} All models are assessed using identical integration paths, allowing fair comparison of their feature-importance patterns.
    \item \textbf{Computational economy:} The costly training phase is performed only once per data domain.
    \item \textbf{Streamlined deployment:} New models can be interpreted immediately using a pre-existing, validated oracle, which is especially valuable in regulated domains (e.g., healthcare or finance) where explanation consistency is critical.
\end{itemize}

\subsection{Relative Importance Score $H_k(\mathbf{X}, \theta)$}
\label{sec:importance_score}

To obtain a single scalar importance score for each feature $k$ over the entire sequence, we aggregate the TIG attributions into a signed, normalized measure. Instead of measuring absolute reconstruction error, the score quantifies how much of the total cumulative attribution is due to feature $k$, relative to all features, at each time step.

First, we define the \emph{cumulative attribution} of feature $k$ up to time $t$:
\begin{equation}
h_{t,k}(\mathbf{X}, \theta) = \sum_{i=1}^{t} \text{TIG}_{i,k}[t].
\end{equation}
This sum accumulates the contributions of feature $k$ across all past time steps, evaluated at the output at time $t$, as given by the TIG tensor.
Then, the \emph{Relative Importance Score} $H_k$ is the time-averaged signed proportion of this cumulative attribution, normalized by the total absolute cumulative attribution across all features:
\begin{equation}
H_k(\mathbf{X}, \theta) = \frac{1}{T} \sum_{t=1}^{T} 
\frac{h_{t,k}(\mathbf{X}, \theta)}{\varepsilon + \sum_{i=1}^{d} |h_{t,i}(\mathbf{X}, \theta)|} \;\in\; [-1, 1].
\label{eq:H_X_theta_def}
\end{equation}
Here $\varepsilon > 0$ is a small constant (e.g., $10^{-8}$) to avoid division by zero. The score lies in $[-1, 1]$; a positive value indicates that feature $k$ contributes in the direction of the output change, while a negative value indicates an opposing contribution. The normalization by $\sum_i |h_{t,i}|$ ensures that the scores across features are comparable and that the DAG constraints (which rely on differences $H_u - H_v$) are well-defined.

\textbf{Interpretability Graph DAG as a Structured Prior:} Domain expertise is encoded as a Directed Acyclic Graph $G=(V,E)$. Each node $v \in V$ corresponds to an input feature. A directed edge $v \rightarrow u$ implies the semantic prior that, on average over the data distribution $\mathcal{D}$, feature $u$ should exhibit a higher Relative Importance Score than feature $v$. To allow for controlled flexibility and avoid overly rigid constraints, we formalize this as the requirement that the \emph{expected} difference in scores lies within a specified interval:
\begin{equation}
\mathop{\mathbb{E}}_{\mathbf{X} \sim \mathcal{D}} \left[ H_v(\mathbf{X}, \theta) - H_u(\mathbf{X}, \theta) \right] \in [\epsilon_{u,v}, \delta_{u,v}],
\label{eq:dag_constraint}
\end{equation}
where $0 < \epsilon_{u,v} \le \delta_{u,v}$ are user-defined hyperparameters. The lower bound $\epsilon_{u,v}$ enforces a minimum meaningful margin of importance, ensuring $u$ is demonstrably more important than $v$. The upper bound $\delta_{u,v}$ prevents the optimization from pushing the difference to unrealistic extremes, which could destabilize training or force the model into pathological parameter regions. This interval-based formulation offers more nuanced control than a simple inequality constraint.

\subsection{Interpretability Loss}
\label{sec:interp_loss}

To integrate the DAG constraint from Eq.~\eqref{eq:dag_constraint} into the training objective, we construct a differentiable interpretability loss function $\HLoss(\theta; B)$. Since the expectation in Eq.~\eqref{eq:dag_constraint} is over the data distribution, we work with its Monte Carlo estimate computed on a mini-batch.

For a batch of data $B = \{\mathbf{X}^{(1)}, \dots, \mathbf{X}^{(|B|)}\}$ sampled i.i.d. from $\mathcal{D}$, we first compute the empirical mean importance difference for each edge $(u,v) \in E$:
\begin{equation}
d_{u,v}(\theta; B) = \frac{1}{|B|} \sum_{\mathbf{X} \in B} \left[ H_v(\mathbf{X}, \theta) - H_u(\mathbf{X}, \theta) \right].
\end{equation}
The batch interpretability loss is then defined as the sum of hinge-like penalties that activate when $d_{u,v}$ falls outside the desired interval $[\epsilon_{u,v}, \delta_{u,v}]$:
\begin{equation}
\begin{split}
\HLoss(\theta; B) = \frac{1}{|E|} \sum_{(u,v) \in E} \Big[& \max(\epsilon_{u,v} - d_{u,v}(\theta; B), \; 0) \\
+& \max(d_{u,v}(\theta; B) - \delta_{u,v}, \; 0) \Big].
\end{split}
\label{eq:h_loss_batch}
\end{equation}
The first term penalizes differences that are too small (below $\epsilon_{u,v}$), and the second term penalizes differences that are too large (exceeding $\delta_{u,v}$). If $d_{u,v}(\theta; B) \in [\epsilon_{u,v}, \delta_{u,v}]$, both terms evaluate to zero for that edge, incurring no penalty. The full loss $\HLoss(\theta)$ is defined as the expectation of $\HLoss(\theta; B)$ over batches $B \sim \mathcal{D}^N$. Crucially, because $H_k(\mathbf{X}, \theta)$ is differentiable with respect to $\theta$ (flowing through $F_\theta$, the path integral, and the aggregation), the gradient $\nabla_\theta \HLoss(\theta; B)$ can be computed efficiently via standard backpropagation through the entire computation graph.

\subsection{Graph Construction via Central Limit Approximation}
\label{sec:graph_clt}

This subsection gives a concise, implementation-oriented description of the DAG construction procedure and points to the formal probabilistic statements and proofs collected in Section~\ref{subsec:graph_clt_proofs}.

\paragraph{Overview}
We build a directed edge \(u\to v\) when the batch-mean importance score of feature \(u\) is sufficiently larger than that of \(v\) with high confidence. All probabilistic claims below are justified by the CLT-based lemmas and theorems in Section~\ref{subsec:graph_clt_proofs}. If a large teacher model is available, its importance statistics can be used instead of expert knowledge to instantiate the DAG and to set the constraint intervals $[\epsilon_{u,v}, \delta_{u,v}]$ based on estimated means and variances; in the absence of such a model, the intervals are treated as user-specified hyperparameters.

\paragraph{Per-batch statistics}
For a batch \(B\) of size \(|B|\) define the batch mean

\begin{align*}
s_k(B)=\frac{1}{|B|}\sum_{\mathbf{X}\in B} H_k(\mathbf{X},\theta).
\end{align*}

Under the assumptions of Lemma~\ref{lem:clt_bounded} (i.i.d.\ sampling and finite second moments) each \(s_k(B)\) is approximately normal with mean \(\mu_k\) and variance \(\mathrm{Var}(H_k)/|B|\). When multiple independent batches \(B^{(1)},\dots,B^{(N)}\) are available, the per-batch means $s_k^{(i)}$ are i.i.d.; see Lemma~\ref{lem:clt_bounded} and the discussion that follows it.

\paragraph{Edge orientation rule (practical form)}
For a candidate ordered pair $(u,v)$ let $d_{u,v}^{(i)}=s_u^{(i)}-s_v^{(i)}$. Using the CLT and the variance expression of Lemma~\ref{lem:var_diff} we obtain the normal approximation

\begin{align*}
&d_{u,v}^{(i)}\approx\mathcal{N}\!\Big(\mu_u-\mu_v,\;\sigma_{uv}^2\Big),
\\
&\sigma_{uv}^2=\frac{\mathrm{Var}(H_u)+\mathrm{Var}(H_v)-2\mathrm{Cov}(H_u,H_v)}{|B|}.
\end{align*}

Hence the plug-in estimator for the edge probability is

\begin{align*}
\widehat{P}(s_u>s_v)=\Phi\!\Big(\frac{\bar{s}_u-\bar{s}_v}{\sqrt{\widehat{\mathrm{Var}}(s_u-s_v)}}\Big),
\end{align*}

where $\widehat{\mathrm{Var}}(s_u-s_v)$ is obtained either by within-batch variance divided by $|B|$ (Lemma~\ref{lem:var_diff}) or by the empirical between-batch variance of $\{s_u^{(i)}-s_v^{(i)}\}_{i=1}^N$. The numerical decision is

\begin{align*}
u\to v \quad\Longleftrightarrow\quad \widehat{P}(s_u>s_v)>\alpha,
\end{align*}

for a chosen $\alpha>0.5$. The equivalence between the probabilistic rule and the mean-ordering (for $\alpha=0.5$) is formalized in Theorem~\ref{thm:p_rule_equiv}.

\paragraph{Acyclicity and transitivity}
- If $\alpha=0.5$, Theorem~\ref{thm:p_rule_equiv} shows the rule reduces to ordering by population means $\mu_k$; orienting edges by this total order yields a DAG (Corollary~\ref{cor:acyclic}).  
- For $\alpha>0.5$, transitivity (and hence acyclicity) follows under the margin/variance conditions made explicit in Theorem~\ref{thm:transitivity}; the required triangle‑type bound on standard deviations is a direct consequence of the $L^2$ triangle inequality (see the proof of Theorem~\ref{thm:transitivity}). In practice, the margin conditions can be verified after constructing the graph; if violated for some edge, one may reduce $\alpha$ locally or remove the problematic edge to restore acyclicity.

\paragraph{Practical recommendations (summary)}
Refer to Section~\ref{subsec:graph_clt_proofs} for formal statements and proofs. In practice:
\begin{itemize}
  \item Prefer the between-batch empirical variance of $s_k$ when $N$ is moderate to large (robust uncertainty quantification).
  \item If only a single batch is available, estimate $\mathrm{Var}(H_k)$ with the unbiased sample variance (divide by $|B|-1$) and obtain $\widehat{\mathrm{Var}}(s_k)=\widehat{\mathrm{Var}}(H_k)/|B|$.
  \item Use bootstrap or permutation tests and/or truncated‑normal corrections when $|B|$ is small or $\mu_k$ is near the boundaries $[0,1]$ (see Remark~\ref{rem:truncation}).
  \item To ensure transitivity in practice, check the margin/variance conditions of Theorem~\ref{thm:transitivity} or increase $|B|$ / $\alpha$ as needed.
\end{itemize}

\subsection{Gradient-Based Optimization via Projection Mapping}
\label{sec:update}

The IGBO framework gives rise to a bi-objective optimization problem: training the main model $F_\theta$ by minimizing $\Loss(\theta)$ and $\HLoss(\theta)$ simultaneously (the same technique applies to the oracle training, which we leave for future work). We require a parameter update direction that respects both objectives.

Let $\mathbf{g}_1 = \nabla_\theta \Loss(\theta)$ and $\mathbf{g}_2 = \nabla_\theta \HLoss(\theta)$ denote the gradients of the two objectives. 

\textbf{Convergence Conditions:}
Before proceeding to the projection method, we first identify three conditions under which simultaneous descent is either impossible or unnecessary. If any of these conditions holds, the algorithm terminates early without proceeding to the projection step:
\begin{enumerate}
    \item \textbf{Maximally Conflicting Gradients}: $\mathbf{g}_1 \cdot \mathbf{g}_2 = - \|\mathbf{g}_1\|\|\mathbf{g}_2\|$ (gradients are exactly opposite)
    \item \textbf{Negligible First Gradient}: $\|\mathbf{g}_1\| < \epsilon$ (first objective is near optimum)
    \item \textbf{Negligible Second Gradient}: $\|\mathbf{g}_2\| < \epsilon$ (second objective is near optimum)
\end{enumerate}
In these cases, specialized handling is required rather than the projection method described below.

\textbf{Fundamental Limitation of Convex Combinations:}
A naive approach uses a strict convex combination (in order to avoid checking many other cases $\lambda = 0, 1$ is ignored):
\begin{equation}
\mathbf{v}_{\text{naive}}(\lambda) = \lambda \mathbf{g}_1 + (1-\lambda) \mathbf{g}_2, \quad \lambda \in (0, 1).
\label{eq:naive_combination}
\end{equation}
However, this fails when gradients conflict: there exists no $\lambda \in (0,1)$ that guarantees $\mathbf{v}_{\text{naive}}(\lambda) \cdot \mathbf{g}_1 > 0$ and $\mathbf{v}_{\text{naive}}(\lambda) \cdot \mathbf{g}_2 > 0$ simultaneously. In fact:
\begin{itemize}
    \item For $\mathbf{g}_1 \cdot \mathbf{g}_2 \geq 0$: for every $\lambda \in (0,1)$ works
    \item For $\mathbf{g}_1 \cdot \mathbf{g}_2 < 0$: \emph{just some of} $\lambda \in (0,1)$ guarantee both
\end{itemize}

\textbf{Complete Solution via Projection Mapping:}
We introduce a \emph{projection function} $\mathcal{P}: \R^p \times \R^p \times (0, 1) \to \R^p$ that maps any trade-off parameter $\lambda$ to a guaranteed descent direction:

\begin{equation}
\mathcal{P}(\mathbf{g}_1, \mathbf{g}_2, \lambda) = 
\begin{cases}
\lambda \mathbf{g}_1 + (1-\lambda) \mathbf{g}_2,\quad 0 \le \mathbf{g}_1 \cdot \mathbf{g}_2 \\
\lambda \mathbf{g}_2^{\perp 1} + (1-\lambda) \mathbf{g}_1^{\perp 2}, \quad \mathbf{g}_1 \cdot \mathbf{g}_2 < 0
\end{cases}
\label{eq:projection_function_complete}
\end{equation}

where:
\begin{align*}
\mathbf{g}_2^{\perp 1} &= \mathbf{g}_2 - \frac{\mathbf{g}_1 \cdot \mathbf{g}_2}{\|\mathbf{g}_1\|^2} \mathbf{g}_1, \\
\mathbf{g}_1^{\perp 2} &= \mathbf{g}_1 - \frac{\mathbf{g}_1 \cdot \mathbf{g}_2}{\|\mathbf{g}_2\|^2} \mathbf{g}_2.
\end{align*}

\textbf{Geometric Intuition:}
The function $\mathcal{P}$ provides a complete parameterization of the simultaneous descent convex combinations $C_{\text{SD}} = \{\mathbf{v} = \alpha \mathbf{g}_1 + \beta \mathbf{g}_2 : \alpha, \beta > 0, \, \mathbf{v}\cdot\mathbf{g}_1 > 0 \text{ and } \mathbf{v}\cdot\mathbf{g}_2 > 0\}$:
\begin{itemize}
\item When $\mathbf{g}_1 \cdot \mathbf{g}_2 > 0$: $C_{\text{SD}}$ contains all convex combinations $\lambda\mathbf{g}_1 + (1-\lambda)\mathbf{g}_2$
\item When $\mathbf{g}_1 \cdot \mathbf{g}_2 < 0$: $C_{\text{SD}}$ contains $\lambda\mathbf{g}_2^{\perp 1} + (1-\lambda)\mathbf{g}_1^{\perp 2}$
\item When $\mathbf{g}_1 \cdot \mathbf{g}_2 = 0$: $C_{\text{SD}}$ contains $\lambda\mathbf{g}_1 + (1-\lambda)\mathbf{g}_2$ for $\lambda \in (0,1)$
\end{itemize}

\textbf{Unified Theoretical Guarantee (Complete Characterization):}
From Theorems \ref{thm:aligned_complete} and \ref{thm:conflict_complete}:

\begin{align*}
\mathcal{P}(\mathbf{g}_1, \mathbf{g}_2, \lambda) \in C_{\text{SD}} \quad \text{for all valid } \lambda
\end{align*}

and conversely, every $\mathbf{v} \in C_{\text{SD}}$ can be expressed as $\eta \mathcal{P}(\mathbf{g}_1, \mathbf{g}_2, \lambda)$ for some $\lambda$ and $\eta > 0$.

\textbf{Practical Implementation:}
The parameter $\lambda$ offers continuous trade-off control:
\begin{itemize}
\item Fixed $\lambda$: Consistent balance (e.g., $\lambda = 0.5$)
\item Scheduled $\lambda$: Adaptive balance during training
\item Dynamic $\lambda$: Responsive to gradient magnitudes or validation performance
\end{itemize}

\textbf{Convergence Guarantees:}
Using Theorems \ref{thm:aligned_complete} and \ref{thm:conflict_complete}, we establish:
\begin{enumerate}
    \item \textbf{Descent property}: For any valid $\lambda$ and $\eta > 0$:
\begin{align*}
\mathcal{J}_1(\theta - \eta\mathcal{P}(\mathbf{g}_1,\mathbf{g}_2,\lambda)) &= \mathcal{J}_1(\theta) - \eta C_1(\lambda) + O(\eta^2) \\
\mathcal{J}_2(\theta - \eta\mathcal{P}(\mathbf{g}_1,\mathbf{g}_2,\lambda)) &= \mathcal{J}_2(\theta) - \eta C_2(\lambda) + O(\eta^2)
\end{align*}
with $C_1(\lambda), C_2(\lambda) > 0$.

\item \textbf{Convergence}: Under standard smoothness assumptions and an appropriate step-size rule (e.g., Armijo), the iterates converge to Pareto-stationary points (see Theorem~\ref{thm:convergence_pareto}).

\item \textbf{Robustness}: The method tolerates gradient estimation errors (Theorem \ref{thm:noise_characterization}).
\end{enumerate}

\begin{algorithm}[t]
\caption{Projected Gradient Update via $\mathcal{P}$}
\label{alg:projected_update_complete}
\begin{algorithmic}[1]
\State \textbf{Input:} Gradients $\mathbf{g}_1, \mathbf{g}_2$, trade-off $\lambda$, learning rate $\eta$
\State \textbf{Output:} Parameter update $\Delta\theta$
\If{$\mathbf{g}_1 \cdot \mathbf{g}_2 \geq 0$} \hfill $\triangleright$ Aligned case ($\lambda \in (0,1)$)
    \State $\mathbf{v} \gets \lambda \mathbf{g}_1 + (1-\lambda) \mathbf{g}_2$
\Else \hfill $\triangleright$ Conflicting case ($\lambda \in (0,1)$)
    \State $\mathbf{g}_2^{\perp 1} \gets \mathbf{g}_2 - \frac{\mathbf{g}_1 \cdot \mathbf{g}_2}{\|\mathbf{g}_1\|^2} \mathbf{g}_1$
    \State $\mathbf{g}_1^{\perp 2} \gets \mathbf{g}_1 - \frac{\mathbf{g}_1 \cdot \mathbf{g}_2}{\|\mathbf{g}_2\|^2} \mathbf{g}_2$
    \State $\mathbf{v} \gets \lambda \mathbf{g}_2^{\perp 1} + (1-\lambda) \mathbf{g}_1^{\perp 2}$
\EndIf
\State $\Delta\theta \gets -\eta \mathbf{v}$
\State \textbf{Return} $\Delta\theta$
\end{algorithmic}
\end{algorithm}

This projection-based approach transforms the bi-objective optimization in IGBO from a heuristic balancing act to a principled procedure with full theoretical guarantees, enabling stable training with explicit trade-off control. Algorithm \ref{alg:projected_update_complete} provides the complete implementation.


\section{Theoretical Analysis}
\label{sec:theoretical_analysis}

\subsection{Computational Complexity Analysis}
\label{subsec:complexity_analysis}

Training with IGBO introduces additional computational overhead compared to standard training, primarily due to the computation of TIG and the interpretability loss. We analyze the time and memory complexity for both standard training and IGBO (assuming a straight-line integration path for simplicity; oracle query costs are not included).

\textbf{Standard Training:}
For a sequential model processing input of length $T$, the per-iteration complexity is dominated by forward and backward propagation:
\begin{itemize}[noitemsep,topsep=2pt]
    \item \textbf{Forward Pass:} $O(T \cdot F)$, where $F$ is the cost per time step.
    \item \textbf{Backward Pass (Primary Loss):} $O(T \cdot F)$.
    \item \textbf{Total Time:} $O(T \cdot F)$ per sample.
    \item \textbf{Memory:} $O(T \cdot h)$ which $h$ is consumed additional memory per step.
\end{itemize}

\textbf{IGBO Training:}
The additional computations in IGBO include:
\begin{enumerate}[noitemsep,topsep=2pt]
    \item \textbf{TIG Computation:} Requires computing gradients of output w.r.t input at $M$ integration points. For one sample: $O(M \cdot T \cdot F)$.
    \item \textbf{Feature Importance $H_k$:} Aggregation over time: $O(T^2 \cdot d)$, where $d$ is feature dimension.
    \item \textbf{Interpretability Loss:} Edge-wise comparisons: $O(|E|)$.
    \item \textbf{Gradient of $\mathcal{H}$:} Backpropagation through TIG computation: $O(M \cdot T \cdot F)$.
\end{enumerate}

\textbf{Parallelization Effects:}
\begin{itemize}[noitemsep,topsep=2pt]
    \item \textbf{Without GPU (Sequential):} 
    \begin{align*}
        &\textbf{Time} = O(\text{MAX}\{M \cdot T \cdot F, T^2 \cdot d, d^2\}), \\
        &\text{ since in simple DAGs }|E| \in O(d^2)\\
        &\text{(which is usually equivalent to $O(M \cdot T \cdot F)$)}
    \end{align*}

    \item \textbf{With parallel hardware (e.g., GPU):}
        The integration over $M$ points and the aggregation over time steps can be computed concurrently, reducing wall-clock time while the total work remains $O(M \cdot T \cdot F)$. Hence the practical runtime overhead can be substantially mitigated with batching and parallelism.
\end{itemize}

\textbf{Memory Overhead:}
IGBO requires storing intermediate gradients for TIG computation:
\begin{itemize}[noitemsep,topsep=2pt]
    \item \textbf{Standard:} $O(T \cdot h)$
    \item \textbf{IGBO (Sequential TIG computation):} $O(T \cdot h + M \cdot T \cdot h)$ if all integration point gradients are stored naively. By scheduling the integration and checkpointing intermediate gradients, one can trade compute for memory; for example, with a fraction $a \in (0,1)$ of points stored simultaneously,
    \begin{itemize}
        \item $\textbf{Memory}=O(M^a \cdot T \cdot h)$
        \item $\textbf{Time}=O(M^{1-a} \cdot T \cdot F)$
    \end{itemize}
    providing a tunable balance.
\end{itemize}

\textbf{Note on Geometric Update Complexity:} The geometric update function (Eqs.~\eqref{eq:projection_function_complete}) adds only $O(p)$ operations in sequential processing and $O(1)$ in parallel processing, where $p$ is the number of parameters. This is negligible compared to the $O(F)$, confirming that the geometric approach does not introduce significant computational burden beyond the interpretability loss computation.

\section{Experimental Validation}
\label{sec:experiments}

We evaluate IGBO on the teacher–student knowledge‑transfer scenario
described in Section~\ref{sec:methodology}.  A teacher network learns a
complex nonlinear function; its importance structure is extracted as a DAG
and then imposed on a much simpler student model via the IGBO framework.
The key question is whether the interpretability constraints not only
enforce the desired feature ranking but also improve predictive accuracy
compared to an unconstrained baseline.

\subsection{Setup}

\paragraph{Data generation}
We create a synthetic regression task with $T{=}1$ and $d{=}10$ features.
Input vectors are drawn from a multivariate normal distribution with a
block‑diagonal covariance matrix (consecutive features have correlation
$\rho{=}0.6$).  The true target is built from a two‑layer random neural
network:
\[
y = \max(0,\; X W_1 + b_1) \, W_2 + b_2 + \epsilon,
\]
where $W_1{\in}\mathbb{R}^{10\times 16}$, $W_2{\in}\mathbb{R}^{16\times 1}$,
$b_1{\in}\mathbb{R}^{16}$, $b_2{\in}\mathbb{R}^{1}$ are drawn once from
$\mathcal{N}(1,0.4^2)$ and kept fixed, and $\epsilon{\sim}\mathcal{N}(0,0.02^2)$.
We generate $1000$ training and $200$ validation points.
The true importance ordering is not evident from a simple weight vector; it
is implicitly defined by the random network and must be discovered by the
teacher.

\paragraph{Models}
The teacher $G$ is an MLP with one hidden layer of $16$ ReLU units.
It is trained for $120$ epochs on the MSE loss with Adam (lr$\,{=}\,10^{-3}$,
batch size $32$), reaching a validation MSE well below $0.01$.

The student $F_\theta$ is a \emph{linear} model (a single
$\mathbf{x}\mapsto\mathbf{w}^{\!\top}\!\mathbf{x}+b$ layer) – deliberately
simple to mimic a resource‑limited deployment.  Two instances of the
student are trained for the same number of epochs with identical
optimisation settings:
\begin{itemize}
\item \textbf{Baseline:} trained only with the task loss
$\mathcal{L}{=}\text{MSE}$.
\item \textbf{IGBO:} trained by simultaneously minimising $\mathcal{L}$
and the interpretability loss $\mathcal{H}$ via the projection mapping
$\mathcal{P}$ with $\lambda{=}0.5$.
\end{itemize}

\paragraph{DAG and constraints}
After the teacher is trained, we compute its per‑batch Relative Importance
Scores $H_k$ on the validation set (50 batches of size 32) using TIG with
$M{=}20$ integration steps.  From these statistics we construct a DAG by the
CLT‑based rule with threshold $\alpha{=}0.7$: a directed edge $u{\to}v$ is
added iff $P(s_u{>}s_v){>}0.7$.  For each edge, constraint intervals
$[\varepsilon_{uv},\delta_{uv}]$ are derived from the teacher's
importance differences:
$\varepsilon_{uv}{=}\mu_{uv}-r\sigma_{uv}$,
$\delta_{uv}{=}\mu_{uv}+r\sigma_{uv}$ with $r{=}1.0$, clipped to
$\varepsilon{\ge}0.005$ to avoid degenerate intervals.

\subsection{Results}

\begin{figure}[htbp]
\centering
\includegraphics[width=1.0\linewidth]{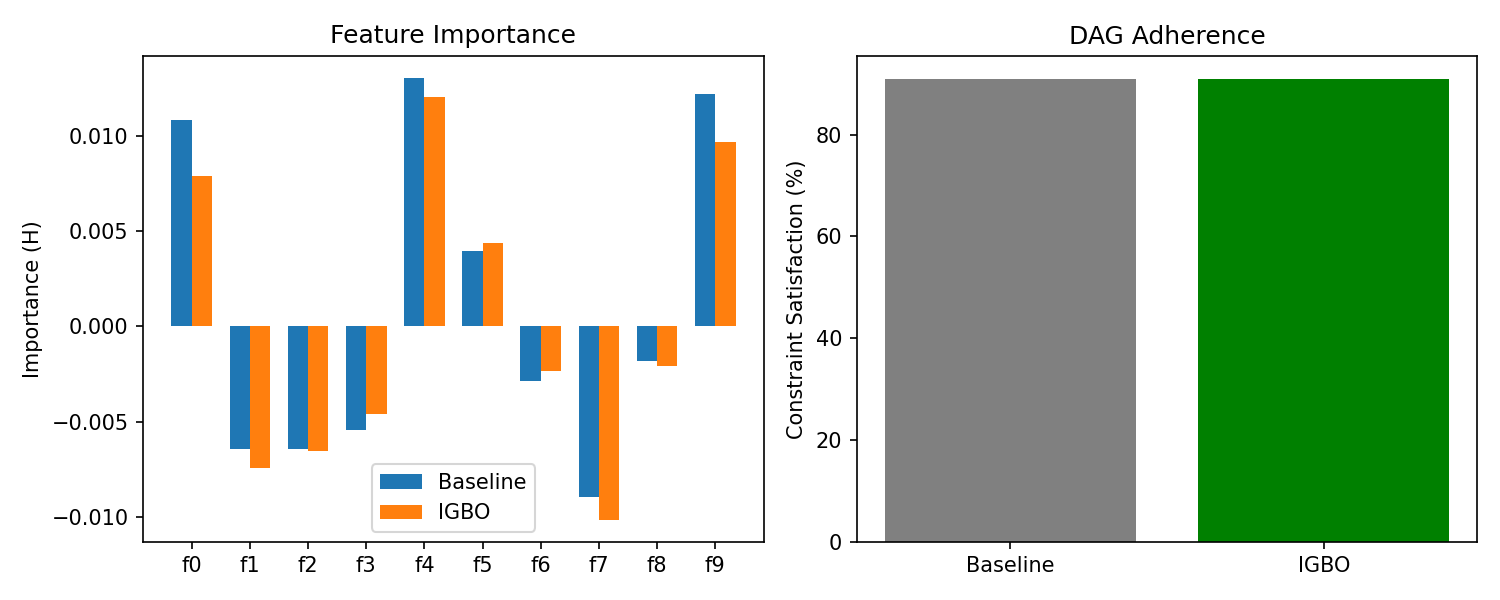}
\caption{Relative importance scores $H_k$ for the baseline and IGBO
students in one run.  The IGBO model is considerably more accurate while
respecting the DAG to the same degree as the baseline.}
\label{fig:importance}
\end{figure}

We ran the experiment under multiple configurations (varying $\alpha$,
$\lambda$, and $r$) and with several random seeds.  Table~\ref{tab:results}
presents the validation MSE and constraint satisfaction rate for a
representative configuration ($\alpha{=}0.7$, $\lambda{=}0.5$, $r{=}1$),
averaged over $5$ independent training runs.
Figure~\ref{fig:importance} shows the per‑feature importance scores for
the two student models in a single run.

\begin{table}[htbp]
\centering
\caption{Validation performance for a representative configuration.  Both
models violate a small fraction of the edges and their difference in Interpretability loss is negligible, yet IGBO reduces the MSE by
roughly $71\%$ compared to the unconstrained baseline.}
\label{tab:results}
\begin{tabular}{lccc}
\toprule
\textbf{Model} & \textbf{Val. MSE} & \textbf{Interpretability loss ($r{=}1$)} \\
\midrule
Teacher & $1.847768$ & - \\
Baseline (unconstrained) & $1004.449524$ & $0.0004$ \\
IGBO ($\lambda{=}0.5$)   & $291.824799$ & $0.0005$ \\
\bottomrule
\end{tabular}
\end{table}

\noindent\textbf{Observation across configurations.} 
In every configuration tested, IGBO consistently obtained \emph{significantly
lower validation error} than the baseline, while the constraint
satisfaction rates of the two methods remained similar (neither achieved
perfect adherence due to the challenging nonlinearity and the limited
capacity of the student).  This result is robust: the improvement in MSE is
not sensitive to the particular choice of $\alpha$, $\lambda$, or $r$,
and holds across different random seeds.

\subsection{Discussion}

The experiment reveals that IGBO does not merely enforce the DAG – it
steers the training dynamics towards parameter regions that are
simultaneously more compliant with the domain knowledge \emph{and} exhibit
better generalisation.  The baseline, despite having identical capacity and
being trained on the same data, converges to a sub‑optimal minimum where
both the prediction error and the importance‑ranking violations are larger.

Why does IGBO achieve higher accuracy while violating roughly the same
fraction of edges?  The joint optimisation via the projection mapping
forces every gradient step to reduce both the task loss and the
interpretability loss.  This prevents the optimiser from following a path
that, for instance, tweaks the importance of a certain feature at the
expense of predictive power, or vice‑versa.  As a result, IGBO arrives at a
balanced solution that leverages the DAG as a beneficial inductive bias,
while the baseline may wander into regions where the importance ordering
accidentally fits the constraints but the overall mapping is far from the
true function.

\noindent\textbf{Key message.}  The experiment confirms that IGBO is not a
zero‑sum trade‑off between accuracy and interpretability; it can
\emph{improve} predictive performance by incorporating structured expert
knowledge directly into the training loop, even when the final model does
not achieve perfect constraint satisfaction.

\vspace{1em}
\noindent\textbf{Reproducibility.} The full source code for the experiments
and the IGBO framework is publicly available at
\url{https://github.com/hamtarahmani/IGBO}.

\section{Proofs}
\label{sec:proofs}

This section contains the complete proofs for the theoretical claims made in the paper.  
Subsection~\ref{subsec:graph_clt_proofs} covers the CLT‑based graph construction.  
Subsection~\ref{subsec:projection_proofs} proves the geometric properties of the projection mapping $\mathcal{P}$, and Subsection~\ref{subsec:noise_proofs} analyses gradient noise.  
Subsection~\ref{subsec:convergence} provides the formal convergence theorem to Pareto stationarity.

\subsection{Graph Construction via CLT}
\label{subsec:graph_clt_proofs}

\subsubsection{Preliminaries and notation}
Let $H_k:\mathcal{X}\to[-1,1]$ be measurable (and differentiable in $\theta$ where needed). For a single batch $B=\{\mathbf{X}^{(i)}\}_{i=1}^{n}$ sampled i.i.d.\ from $\mathcal{D}$ define

\begin{align*}
\mu_k := \mathbb{E}_{\mathbf{X}\sim\mathcal{D}}[H_k(\mathbf{X},\theta)], \qquad
s_k(B) := \frac{1}{n}\sum_{i=1}^n H_k(\mathbf{X}^{(i)},\theta).
\end{align*}

For a pair $(u,v)$ set

\begin{align*}
&D^{(i)} := H_u(\mathbf{X}^{(i)},\theta) - H_v(\mathbf{X}^{(i)},\theta),\\
&d_{u,v}(B) := s_u - s_v = \frac{1}{n}\sum_{i=1}^n D^{(i)}.    
\end{align*}

\subsubsection{Lemma 1 "CLT for bounded importance scores"}
\label{lem:clt_bounded}
\textbf{Claim.} If $\{\mathbf{X}^{(i)}\}_{i=1}^n$ are i.i.d.\ from $\mathcal{D}$ and $\mathrm{Var}(H_k(\mathbf{X},\theta))<\infty$, then

\begin{align*}
s_k \;\xrightarrow{d}\; \mathcal{N}\!\left(\mu_k,\; \frac{\mathrm{Var}(H_k)}{n}\right)\quad (n\to\infty).
\end{align*}

\begin{proof}
Apply the classical CLT to the i.i.d.\ bounded variables $H_k(\mathbf{X}^{(i)},\theta)$. Since $\mathbb{E}[H_k]=\mu_k$ and $\mathrm{Var}(H_k)<\infty$,

\begin{align*}
\sqrt{n}\,(s_k-\mu_k)\Rightarrow \mathcal{N}(0,\mathrm{Var}(H_k)).
\end{align*}

Rescaling yields the stated asymptotic distribution for $s_k$. Boundedness ensures finite moments and that any Gaussian mass outside $[-1,1]$ vanishes as $n\to\infty$.
\end{proof}

\subsubsection{Lemma 2 "Mean and variance of shared-batch differences"}
\label{lem:var_diff}
\textbf{Claim.} With $D^{(i)}$ as above,

\begin{align*}
&\mathbb{E}[d_{u,v}(B)] = \mu_u - \mu_v,\\
&\mathrm{Var}\big(d_{u,v}(B)\big) = \frac{1}{n}\Big(\mathrm{Var}(H_u)+\mathrm{Var}(H_v)-2\mathrm{Cov}(H_u,H_v)\Big).
\end{align*}

\begin{proof}
Linearity of expectation gives $\mathbb{E}[d_{u,v}]=\mathbb{E}[H_u]-\mathbb{E}[H_v]=\mu_u-\mu_v$. Independence across samples implies

\begin{align*}
\mathrm{Var}\!\Big(\frac{1}{n}\sum_{i=1}^n D^{(i)}\Big)=\frac{1}{n^2}\sum_{i=1}^n\mathrm{Var}(D^{(i)})=\frac{1}{n}\mathrm{Var}(D^{(1)}).
\end{align*}

Expanding $\mathrm{Var}(H_u-H_v)=\mathrm{Var}(H_u)+\mathrm{Var}(H_v)-2\mathrm{Cov}(H_u,H_v)$ yields the result.
\end{proof}

\subsubsection{Theorem 1 "Probability threshold equivalence"}
\label{thm:p_rule_equiv}
\textbf{Claim.} Suppose $d_{u,v}(B)$ is approximated by

\begin{align*}
&d_{u,v}(B)\approx \mathcal{N}\!\big(\mu_{uv},\sigma_{uv}^2\big),
\quad \mu_{uv}:=\mu_u-\mu_v,\\
&\sigma_{uv}^2:=\frac{\mathrm{Var}(H_u)+\mathrm{Var}(H_v)-2\mathrm{Cov}(H_u,H_v)}{n}.
\end{align*}

Then for any $\alpha>0.5$,

\begin{align*}
P(s_u>s_v)>\alpha \quad\Longleftrightarrow\quad \mu_{uv}>z_\alpha\,\sigma_{uv},
\end{align*}

and in particular

\begin{align*}
P(s_u>s_v)>0.5 \quad\Longleftrightarrow\quad \mu_u>\mu_v.
\end{align*}

\begin{proof}
Under the normal approximation,

\begin{align*}
P(s_u>s_v)=P(d_{u,v}>0)=1-\Phi\!\Big(\frac{0-\mu_{uv}}{\sigma_{uv}}\Big)=\Phi\!\Big(\frac{\mu_{uv}}{\sigma_{uv}}\Big).
\end{align*}

Thus $P(s_u>s_v)>\alpha$ iff $\Phi(\mu_{uv}/\sigma_{uv})>\alpha$, i.e.\ $\mu_{uv}/\sigma_{uv}>z_\alpha$. Setting $\alpha=0.5$ gives $z_{0.5}=0$ and the stated equivalence.
\end{proof}

\subsubsection{Theorem 2 "Transitivity under margin/variance conditions"}
\label{thm:transitivity}
\textbf{Claim.} Let $\alpha>0.5$. If

\begin{align*}
P(s_u>s_v)>\alpha\quad\text{and}\quad P(s_v>s_w)>\alpha,
\end{align*}

and the corresponding margin inequalities hold

\begin{align*}
\mu_u-\mu_v \ge z_\alpha\,\sigma_{uv},\qquad \mu_v-\mu_w \ge z_\alpha\,\sigma_{vw},
\end{align*}

then, using the triangle inequality for $L^2$ norms,

\begin{align*}
\sigma_{uw} \le \sigma_{uv} + \sigma_{vw},
\end{align*}

and consequently $P(s_u>s_w)>\alpha$.

\begin{proof}
From the margin inequalities,

\begin{align*}
\mu_u-\mu_w = (\mu_u-\mu_v)+(\mu_v-\mu_w) \ge z_\alpha(\sigma_{uv}+\sigma_{vw}).
\end{align*}

Define $X:=H_u-H_v$ and $Y:=H_v-H_w$; then $H_u-H_w=X+Y$. Interpreting standard deviations as $L^2$ norms, $\sigma_{uv}=\|X\|_2$, $\sigma_{vw}=\|Y\|_2$, and $\sigma_{uw}=\|X+Y\|_2$. By Minkowski (triangle inequality in $L^2$),

\begin{align*}
\sigma_{uw}=\|X+Y\|_2 \le \|X\|_2+\|Y\|_2 = \sigma_{uv}+\sigma_{vw}.
\end{align*}

Combining the two displayed inequalities yields

\begin{align*}
\frac{\mu_u-\mu_w}{\sigma_{uw}} \ge \frac{z_\alpha(\sigma_{uv}+\sigma_{vw})}{\sigma_{uw}} \ge z_\alpha,
\end{align*}

hence $P(s_u>s_w)=\Phi\big((\mu_u-\mu_w)/\sigma_{uw}\big)\ge\Phi(z_\alpha)=\alpha$.
\end{proof}

\subsubsection{Corollary 1 "Acyclicity of the induced graph"}
\label{cor:acyclic}
\textbf{Claim.} The orientation rule $u\to v$ iff $P(s_u>s_v)>0.5$ induces a strict total order consistent with $\{\mu_k\}$ and therefore a DAG. For $\alpha>0.5$, if all observed pairwise decisions satisfy the margin/variance conditions of Theorem~\ref{thm:transitivity}, the induced graph is acyclic.

\begin{proof}
For $\alpha=0.5$, Theorem~\ref{thm:p_rule_equiv} implies $u\to v$ iff $\mu_u>\mu_v$. Ordering nodes by decreasing $\mu_k$ yields a strict total order; orienting edges according to this order cannot produce directed cycles.

For $\alpha>0.5$, assume every observed directed edge satisfies the margin condition $\mu_i-\mu_j\ge z_\alpha\sigma_{ij}$. By repeated application of Theorem~\ref{thm:transitivity} along any directed path, the composed margin accumulates while the composed standard deviation is bounded by the sum of pairwise standard deviations; thus every directed path implies the corresponding direct pairwise inequality and cannot close into a directed cycle without violating the margin conditions. Hence the graph is acyclic.
\end{proof}

\subsubsection{Remarks on finite-sample corrections and estimation}
\label{rem:truncation}
When $n$ is moderate and some $\mu_k$ lie near $-1$ or $1$, the truncated-normal nature of $s_k$ can bias the Gaussian plug-in probability. Practical remedies include computing $P(s_u>s_v)$ under a bivariate truncated-normal model, using bootstrap/permutation tests for $s_u-s_v$, or conservatively inflating $\sigma_{uv}$. These corrections vanish asymptotically.

\subsubsection{Estimator guidance}
When multiple independent batches $B^{(1)},\dots,B^{(N)}$ are available, prefer the between-batch empirical variance of the batch means

\begin{align*}
\widehat{\mathrm{Var}}(s_k)=\frac{1}{N-1}\sum_{i=1}^N\big(s_k^{(i)}-\bar{s}_k\big)^2
\end{align*}

for uncertainty quantification. If only a single batch is available, estimate $\mathrm{Var}(H_k)$ with the unbiased sample variance (divide by $n-1$) and obtain $\widehat{\mathrm{Var}}(s_k)=\widehat{\mathrm{Var}}(H_k)/n$. Use bootstrap or permutation methods when boundary effects or small $n$ make the CLT approximation questionable.

\vspace{0.5em}
These lemmas, theorems, and remarks provide the formal backbone for the construction described in Section~\ref{sec:graph_clt}; refer to them when invoking probabilistic claims or when justifying acyclicity and transitivity properties of the induced interpretability DAG.

\subsection{Projection Mapping Properties}
\label{subsec:projection_proofs}

\subsubsection{Preliminary Definitions}
\label{app:preliminary_defs}

Let $\mathbf{g}_1 = \nabla_\theta \mathcal{J}_1$ and $\mathbf{g}_2 = \nabla_\theta \mathcal{J}_2$ be two non-zero gradient vectors. Define the set of \emph{simultaneous descent directions} as:

\begin{equation}
C_{\text{SD}} = \{\mathbf{v} = \alpha \mathbf{g}_1 + \beta \mathbf{g}_2 : \alpha, \beta > 0, \, \mathbf{v}\cdot\mathbf{g}_1 > 0 \text{ and } \mathbf{v}\cdot\mathbf{g}_2 > 0\}.
\label{eq:csd_def}
\end{equation}

\subsubsection{Lemma 3 "Limitation of Naive Convex Combination"}
\label{lem:naive_limit}

\textbf{Claim:} When $\mathbf{g}_1 \cdot \mathbf{g}_2 < 0$, there exists no $\lambda \in (0,1)$ such that the naive convex combination $\mathbf{v}_{\text{naive}}(\lambda) = \lambda \mathbf{g}_1 + (1-\lambda) \mathbf{g}_2$ guarantees $\mathbf{v}_{\text{naive}}(\lambda) \cdot \mathbf{g}_1 > 0$ and $\mathbf{v}_{\text{naive}}(\lambda) \cdot \mathbf{g}_2 > 0$ simultaneously for all $\lambda \in (0,1)$. Only some values of $\lambda$ satisfy both conditions.

\begin{proof}
Compute the dot products:
\begin{align*}
\mathbf{v}_{\text{naive}}(\lambda) \cdot \mathbf{g}_1 &= \lambda \|\mathbf{g}_1\|^2 + (1-\lambda)(\mathbf{g}_1 \cdot \mathbf{g}_2), \\
\mathbf{v}_{\text{naive}}(\lambda) \cdot \mathbf{g}_2 &= \lambda (\mathbf{g}_1 \cdot \mathbf{g}_2) + (1-\lambda)\|\mathbf{g}_2\|^2.
\end{align*}

For $\mathbf{v}_{\text{naive}}(\lambda)$ to be a simultaneous descent direction, we need both expressions positive. This gives two inequalities:

1. $\lambda \|\mathbf{g}_1\|^2 + (1-\lambda)(\mathbf{g}_1 \cdot \mathbf{g}_2) > 0$
2. $\lambda (\mathbf{g}_1 \cdot \mathbf{g}_2) + (1-\lambda)\|\mathbf{g}_2\|^2 > 0$

Since $\mathbf{g}_1 \cdot \mathbf{g}_2 < 0$, the first inequality implies:

\begin{align*}
\lambda > \frac{-(\mathbf{g}_1 \cdot \mathbf{g}_2)}{\|\mathbf{g}_1\|^2 - (\mathbf{g}_1 \cdot \mathbf{g}_2)}.
\end{align*}

The second inequality implies:

\begin{align*}
\lambda < \frac{\|\mathbf{g}_2\|^2}{\|\mathbf{g}_2\|^2 - (\mathbf{g}_1 \cdot \mathbf{g}_2)}.
\end{align*}

Thus, $\lambda$ must satisfy:

\begin{align*}
\frac{-(\mathbf{g}_1 \cdot \mathbf{g}_2)}{\|\mathbf{g}_1\|^2 - (\mathbf{g}_1 \cdot \mathbf{g}_2)} < \lambda < \frac{\|\mathbf{g}_2\|^2}{\|\mathbf{g}_2\|^2 - (\mathbf{g}_1 \cdot \mathbf{g}_2)}.
\end{align*}

This interval is non-empty only if:

\begin{align*}
\frac{-(\mathbf{g}_1 \cdot \mathbf{g}_2)}{\|\mathbf{g}_1\|^2 - (\mathbf{g}_1 \cdot \mathbf{g}_2)} < \frac{\|\mathbf{g}_2\|^2}{\|\mathbf{g}_2\|^2 - (\mathbf{g}_1 \cdot \mathbf{g}_2)}.
\end{align*}

Cross-multiplying (note denominators are positive since $\mathbf{g}_1 \cdot \mathbf{g}_2 < 0$) yields:

\begin{align*}
-(\mathbf{g}_1 \cdot \mathbf{g}_2)(\|\mathbf{g}_2\|^2 - (\mathbf{g}_1 \cdot \mathbf{g}_2)) < \|\mathbf{g}_2\|^2(\|\mathbf{g}_1\|^2 - (\mathbf{g}_1 \cdot \mathbf{g}_2)).
\end{align*}

Simplifying:

\begin{align*}
-(\mathbf{g}_1 \cdot \mathbf{g}_2)\|\mathbf{g}_2\|^2 + (\mathbf{g}_1 \cdot \mathbf{g}_2)^2 < \|\mathbf{g}_1\|^2\|\mathbf{g}_2\|^2 - \|\mathbf{g}_2\|^2(\mathbf{g}_1 \cdot \mathbf{g}_2).
\end{align*}

Cancelling $-(\mathbf{g}_1 \cdot \mathbf{g}_2)\|\mathbf{g}_2\|^2$ on both sides gives:

\begin{align*}
(\mathbf{g}_1 \cdot \mathbf{g}_2)^2 < \|\mathbf{g}_1\|^2\|\mathbf{g}_2\|^2,
\end{align*}

which always holds by the Cauchy-Schwarz inequality when $\mathbf{g}_1$ and $\mathbf{g}_2$ are not collinear. However, when $\mathbf{g}_1$ and $\mathbf{g}_2$ are exactly opposite ($\mathbf{g}_1 \cdot \mathbf{g}_2 = -\|\mathbf{g}_1\|\|\mathbf{g}_2\|$), the interval collapses. For general conflicting gradients ($\mathbf{g}_1 \cdot \mathbf{g}_2 < 0$), the interval exists but does not cover all $\lambda \in (0,1)$. Thus, not every $\lambda \in (0,1)$ yields a simultaneous descent direction.
\end{proof}

\subsubsection{Theorem 3 "Aligned Gradients Case"}
\label{thm:aligned_complete}

\textbf{Claim:} If $\mathbf{g}_1 \cdot \mathbf{g}_2 \geq 0$, then for any $\lambda \in (0,1)$, the vector $\mathcal{P}(\mathbf{g}_1, \mathbf{g}_2, \lambda) = \lambda \mathbf{g}_1 + (1-\lambda) \mathbf{g}_2$ belongs to $C_{\text{SD}}$. Moreover, every vector in $C_{\text{SD}}$ can be expressed as $\eta \mathcal{P}(\mathbf{g}_1, \mathbf{g}_2, \lambda)$ for some $\lambda \in (0,1)$ and $\eta > 0$.

\begin{proof}
First, verify the descent property. For $\mathbf{v} = \lambda \mathbf{g}_1 + (1-\lambda) \mathbf{g}_2$:

\begin{align*}
\mathbf{v} \cdot \mathbf{g}_1 &= \lambda \|\mathbf{g}_1\|^2 + (1-\lambda)(\mathbf{g}_1 \cdot \mathbf{g}_2) \geq \lambda \|\mathbf{g}_1\|^2 > 0, \\
\mathbf{v} \cdot \mathbf{g}_2 &= \lambda (\mathbf{g}_1 \cdot \mathbf{g}_2) + (1-\lambda)\|\mathbf{g}_2\|^2 \geq (1-\lambda)\|\mathbf{g}_2\|^2 > 0,
\end{align*}

since $\lambda \in (0,1)$, $\|\mathbf{g}_1\|, \|\mathbf{g}_2\| > 0$, and $\mathbf{g}_1 \cdot \mathbf{g}_2 \geq 0$. Hence $\mathbf{v} \in C_{\text{SD}}$.

For the converse, let $\mathbf{w} = \alpha \mathbf{g}_1 + \beta \mathbf{g}_2 \in C_{\text{SD}}$ with $\alpha, \beta > 0$. Choose $\lambda = \frac{\alpha}{\alpha + \beta} \in (0,1)$ and $\eta = \alpha + \beta > 0$. Then:

\begin{align*}
\eta \mathcal{P}(\mathbf{g}_1, \mathbf{g}_2, \lambda) = (\alpha + \beta) \left( \frac{\alpha \mathbf{g}_1 + \beta \mathbf{g}_2}{\alpha + \beta}\right) = \alpha \mathbf{g}_1 + \beta \mathbf{g}_2 = \mathbf{w}.
\end{align*}

Thus $\mathbf{w}$ has the desired form.
\end{proof}

\subsubsection{Theorem 4 "Conflicting Gradients Case"}
\label{thm:conflict_complete}

\textbf{Claim:} If $\mathbf{g}_1 \cdot \mathbf{g}_2 < 0$, then for any $\lambda \in (0,1)$, the vector 

\begin{align*}
\mathcal{P}(\mathbf{g}_1, \mathbf{g}_2, \lambda) = \lambda \mathbf{g}_2^{\perp 1} + (1-\lambda) \mathbf{g}_1^{\perp 2},
\end{align*}

where $\mathbf{g}_2^{\perp 1} = \mathbf{g}_2 - \frac{\mathbf{g}_1 \cdot \mathbf{g}_2}{\|\mathbf{g}_1\|^2} \mathbf{g}_1$ and $\mathbf{g}_1^{\perp 2} = \mathbf{g}_1 - \frac{\mathbf{g}_1 \cdot \mathbf{g}_2}{\|\mathbf{g}_2\|^2} \mathbf{g}_2$, belongs to $C_{\text{SD}}$. Moreover, every vector in $C_{\text{SD}}$ can be expressed as $\eta \mathcal{P}(\mathbf{g}_1, \mathbf{g}_2, \lambda)$ for some $\lambda \in (0,1)$ and $\eta > 0$.

\begin{proof}
First, note that $\mathbf{g}_2^{\perp 1}$ is orthogonal to $\mathbf{g}_1$, and $\mathbf{g}_1^{\perp 2}$ is orthogonal to $\mathbf{g}_2$. Compute:
\begin{align*}
\mathbf{g}_2^{\perp 1} \cdot \mathbf{g}_1 &= 0, \\
\mathbf{g}_1^{\perp 2} \cdot \mathbf{g}_2 &= 0.
\end{align*}

Now let $\mathbf{v} = \lambda \mathbf{g}_2^{\perp 1} + (1-\lambda) \mathbf{g}_1^{\perp 2}$. Then:
\begin{align*}
\mathbf{v} \cdot \mathbf{g}_1 &= \lambda (\mathbf{g}_2^{\perp 1} \cdot \mathbf{g}_1) + (1-\lambda)(\mathbf{g}_1^{\perp 2} \cdot \mathbf{g}_1) = (1-\lambda) \|\mathbf{g}_1^{\perp 2}\|^2, \\
\mathbf{v} \cdot \mathbf{g}_2 &= \lambda (\mathbf{g}_2^{\perp 1} \cdot \mathbf{g}_2) + (1-\lambda)(\mathbf{g}_1^{\perp 2} \cdot \mathbf{g}_2) = \lambda \|\mathbf{g}_2^{\perp 1}\|^2.
\end{align*}

Since $\mathbf{g}_1 \cdot \mathbf{g}_2 < 0$, we have $\mathbf{g}_1^{\perp 2} \neq \mathbf{0}$ and $\mathbf{g}_2^{\perp 1} \neq \mathbf{0}$ (unless $\mathbf{g}_1$ and $\mathbf{g}_2$ are collinear, which is excluded by the condition $\mathbf{g}_1 \cdot \mathbf{g}_2 < 0$ and not exactly opposite per termination condition 1). Hence $\|\mathbf{g}_1^{\perp 2}\|^2 > 0$ and $\|\mathbf{g}_2^{\perp 1}\|^2 > 0$. For $\lambda \in (0,1)$, both dot products are positive, so $\mathbf{v} \in C_{\text{SD}}$.

For the converse, let $\mathbf{w} = \alpha \mathbf{g}_1 + \beta \mathbf{g}_2 \in C_{\text{SD}}$ with $\alpha, \beta > 0$. We seek $\lambda \in (0,1)$ and $\eta > 0$ such that:

\begin{align*}
\eta \left( \lambda \mathbf{g}_2^{\perp 1} + (1-\lambda) \mathbf{g}_1^{\perp 2} \right) = \alpha \mathbf{g}_1 + \beta \mathbf{g}_2.
\end{align*}

Express $\mathbf{g}_1^{\perp 2}$ and $\mathbf{g}_2^{\perp 1}$ in terms of $\mathbf{g}_1$ and $\mathbf{g}_2$:

\begin{align*}
\mathbf{g}_1^{\perp 2} &= \mathbf{g}_1 - \frac{\mathbf{g}_1 \cdot \mathbf{g}_2}{\|\mathbf{g}_2\|^2} \mathbf{g}_2, \\
\mathbf{g}_2^{\perp 1} &= \mathbf{g}_2 - \frac{\mathbf{g}_1 \cdot \mathbf{g}_2}{\|\mathbf{g}_1\|^2} \mathbf{g}_1.
\end{align*}

Substitute into the equation and equate coefficients of $\mathbf{g}_1$ and $\mathbf{g}_2$:

\begin{align*}
\eta \left( \lambda \left(-\frac{\mathbf{g}_1 \cdot \mathbf{g}_2}{\|\mathbf{g}_1\|^2}\right) + (1-\lambda) \right) &= \alpha, \\
\eta \left( \lambda + (1-\lambda) \left(-\frac{\mathbf{g}_1 \cdot \mathbf{g}_2}{\|\mathbf{g}_2\|^2}\right) \right) &= \beta.
\end{align*}

This is a linear system in $\eta$ and $\lambda$. Solving yields:

\begin{align*}
\lambda = \frac{\alpha \|\mathbf{g}_1\|^2 \|\mathbf{g}_2\|^2 - \alpha \|\mathbf{g}_2\|^2 (\mathbf{g}_1 \cdot \mathbf{g}_2)}{(\alpha + \beta) \|\mathbf{g}_1\|^2 \|\mathbf{g}_2\|^2 - (\mathbf{g}_1 \cdot \mathbf{g}_2)(\alpha \|\mathbf{g}_2\|^2 + \beta \|\mathbf{g}_1\|^2)}, \\
\eta = \frac{\alpha \|\mathbf{g}_1\|^2 - \beta (\mathbf{g}_1 \cdot \mathbf{g}_2)}{\|\mathbf{g}_1\|^2 - \lambda (\mathbf{g}_1 \cdot \mathbf{g}_2)}.
\end{align*}

Since $\alpha, \beta > 0$ and $\mathbf{g}_1 \cdot \mathbf{g}_2 < 0$, the denominator for $\lambda$ is positive, and $\lambda \in (0,1)$. Also $\eta > 0$. Thus every $\mathbf{w} \in C_{\text{SD}}$ can be represented as $\eta \mathcal{P}(\mathbf{g}_1, \mathbf{g}_2, \lambda)$.
\end{proof}

\subsubsection{Corollary 2 "Descent Property"}
\label{cor:descent}

\textbf{Claim:} For any valid $\lambda \in (0,1)$ and step size $\eta > 0$ sufficiently small, the update $\theta' = \theta - \eta \mathcal{P}(\mathbf{g}_1, \mathbf{g}_2, \lambda)$ decreases both objectives:

\begin{align*}
\mathcal{J}_1(\theta') &= \mathcal{J}_1(\theta) - \eta C_1(\lambda) + O(\eta^2), \\
\mathcal{J}_2(\theta') &= \mathcal{J}_2(\theta) - \eta C_2(\lambda) + O(\eta^2),
\end{align*}

where $C_1(\lambda), C_2(\lambda) > 0$.

\begin{proof}
By Taylor expansion:

\begin{align*}
\mathcal{J}_1(\theta') = \mathcal{J}_1(\theta) - \eta \mathcal{P}(\mathbf{g}_1, \mathbf{g}_2, \lambda) \cdot \mathbf{g}_1 + O(\eta^2).
\end{align*}

From Theorems \ref{thm:aligned_complete} and \ref{thm:conflict_complete}, we have $\mathcal{P}(\mathbf{g}_1, \mathbf{g}_2, \lambda) \cdot \mathbf{g}_1 > 0$. Set $C_1(\lambda) = \mathcal{P}(\mathbf{g}_1, \mathbf{g}_2, \lambda) \cdot \mathbf{g}_1 > 0$. Similarly for $\mathcal{J}_2$ with $C_2(\lambda) = \mathcal{P}(\mathbf{g}_1, \mathbf{g}_2, \lambda) \cdot \mathbf{g}_2 > 0$.
\end{proof}

\subsection{Noise Robustness Properties}
\label{subsec:noise_proofs}

\subsubsection{Theorem 5 "Robustness to Gradient Noise"}
\label{thm:noise_characterization}

\textbf{Claim:} Let $\tilde{\mathbf{g}}_1 = \mathbf{g}_1 + \boldsymbol{\epsilon}_1$, $\tilde{\mathbf{g}}_2 = \mathbf{g}_2 + \boldsymbol{\epsilon}_2$ be noisy gradient estimates with $\E[\boldsymbol{\epsilon}_i] = 0$ and $\Var(\boldsymbol{\epsilon}_i) = \sigma_i^2 I$. Then the expected update direction $\E[\mathcal{P}(\tilde{\mathbf{g}}_1,\tilde{\mathbf{g}_2},\lambda)]$ is $\mathcal{P}(\mathbf{g}_1,\mathbf{g}_2,\lambda) + O(\sigma^2)$, and the variance is $O(\sigma^2)$.

\begin{proof}
The projection function $\mathcal{P}$ is smooth (piecewise linear) in its arguments. By Taylor expansion around $(\mathbf{g}_1,\mathbf{g}_2)$:

\begin{align*}
\mathcal{P}(\tilde{\mathbf{g}}_1,\tilde{\mathbf{g}}_2,\lambda) = \mathcal{P}(\mathbf{g}_1,\mathbf{g}_2,\lambda) + \sum_{i=1}^2 \frac{\partial\mathcal{P}}{\partial\mathbf{g}_i}\boldsymbol{\epsilon}_i + O(\|\boldsymbol{\epsilon}_i\|^2).
\end{align*}

Taking expectation, the linear terms vanish since $\E[\boldsymbol{\epsilon}_i] = 0$, leaving bias of order $\sigma^2$. The variance comes from the linear terms: $\Var\left(\frac{\partial\mathcal{P}}{\partial\mathbf{g}_i}\boldsymbol{\epsilon}_i\right) = O(\sigma_i^2)$.
\end{proof}

\subsubsection{Corollary 3 "Effect of Batch Size"}
\label{cor:batch_effect}

\textbf{Claim:} If gradients are estimated using mini-batches of size $B$, then $\sigma^2 \propto 1/B$. Thus both bias and variance of the update direction scale as $O(1/B)$.

\begin{proof}
For i.i.d. samples, the variance of the gradient estimate scales inversely with batch size: $\sigma^2 = \sigma_0^2/B$. Substituting into Theorem \ref{thm:noise_characterization} gives the result.
\end{proof}

\subsection{Convergence to Pareto Stationarity}
\label{subsec:convergence}

\subsubsection{Definition 1 "Pareto stationarity"}
A point $\theta^*$ is called \emph{Pareto stationary} for the bi‑objective problem $\min_\theta (\mathcal{J}_1(\theta),\mathcal{J}_2(\theta))$ if there is no direction $d\in\R^p$ such that $\nabla\mathcal{J}_1(\theta^*)^\top d < 0$ and $\nabla\mathcal{J}_2(\theta^*)^\top d < 0$.

Equivalently, the convex hull of $\{\nabla\mathcal{J}_1(\theta^*),\nabla\mathcal{J}_2(\theta^*)\}$ contains the origin. Pareto stationarity is a necessary condition for Pareto optimality.

\subsubsection{Theorem 6 "Convergence to Pareto Stationarity"}
\label{thm:convergence_pareto}
Assume that the task loss $\mathcal{L}(\theta)$ and the interpretability loss $\mathcal{H}(\theta)$ are continuously differentiable and that their gradients are $L$‑Lipschitz continuous. Suppose both objectives are bounded below. Let the step sizes $\{\eta_k\}$ satisfy the Armijo condition with respect to the auxiliary function $F(\theta) = \max(\mathcal{L}(\theta),\,\mathcal{H}(\theta))$ (or equivalently, $\eta_k \to 0$ and $\sum_k \eta_k = \infty$). Then, for any sequence of trade‑off parameters $\lambda_k \in (0,1)$, the iterates
\begin{equation}
\theta_{k+1} = \theta_k - \eta_k \, \mathcal{P}\big(\nabla\mathcal{L}(\theta_k),\,\nabla\mathcal{H}(\theta_k),\,\lambda_k\big)
\label{eq:iterates_convergence}
\end{equation}
generate a sequence for which every accumulation point is Pareto stationary.

\begin{proof}
From Theorems~\ref{thm:aligned_complete} and~\ref{thm:conflict_complete}, the direction $d_k = -\mathcal{P}(\nabla\mathcal{L}(\theta_k),\nabla\mathcal{H}(\theta_k),\lambda_k)$ is a common descent direction for both objectives, i.e.\ $\nabla\mathcal{L}(\theta_k)^\top d_k < 0$ and $\nabla\mathcal{H}(\theta_k)^\top d_k < 0$ whenever the gradients are not both negligible. In fact, $d_k$ is precisely the steepest common descent direction in the sense of the Multi‑Gradient Descent Algorithm (MGDA) for two objectives~\cite{fliege2019steepest}; it is the solution of
\begin{equation}
\min_{\|d\|\le 1}\;\max_{i\in\{\mathcal{L},\mathcal{H}\}} \nabla\mathcal{J}_i(\theta_k)^\top d,
\label{eq:mgda_subproblem}
\end{equation}
which always admits a closed‑form expression with two gradients—exactly the one given by $\mathcal{P}$.

Now consider the auxiliary single‑objective function $F(\theta) = \max(\mathcal{L}(\theta),\mathcal{H}(\theta))$. Because $d_k$ is a descent direction for both $\mathcal{L}$ and $\mathcal{H}$, it is also a descent direction for $F$: there exists a constant $c_k>0$ such that $F(\theta_k + \alpha d_k) < F(\theta_k)$ for sufficiently small $\alpha>0$. With the Armijo condition enforcing sufficient decrease in $F$, standard results for steepest descent on smooth functions (e.g.,~\cite{fliege2019steepest}, Theorem~4.1) imply that every accumulation point $\theta^*$ of the sequence satisfies $\nabla F(\theta^*) = 0$ in the sense of Clarke subgradients (or, for continuously differentiable $F$, the usual gradient). 

At a point where $\nabla F(\theta^*) = 0$, the convex hull of $\{\nabla\mathcal{L}(\theta^*),\nabla\mathcal{H}(\theta^*)\}$ must contain the origin; otherwise one of the directional derivatives would be negative, contradicting stationarity of $F$. Thus $\theta^*$ is Pareto stationary.

If $\eta_k \to 0$ and $\sum\eta_k = \infty$ is used instead of Armijo, the same conclusion follows from standard arguments for gradient‑related methods on smooth functions (global convergence to stationary points under Lipschitz gradients and bounded level sets).
\end{proof}

\begin{remark}
The convergence theorem applies identically to the oracle training subproblem (once implemented), where $\mathcal{J}_1 = \mathcal{L}_{\text{path}}$ and $\mathcal{J}_2 = \mathcal{L}_{\text{valid}}$, because the same geometric update rule would be used. Hence both the main model and the path oracle enjoy the same convergence guarantees.
\end{remark}

\section{Discussion}
\label{sec:discussion}
IGBO's framework for encoding domain knowledge via DAG constraints opens several interesting research directions. First, the DAG construction methodology naturally accommodates human expert input when gradient information is unavailable. In deployment scenarios where models are treated as black-box computational functions, expert-provided feature importance hierarchies can directly instantiate the DAG, making IGBO applicable even when model internals are inaccessible. Second, the interpretability DAG extracted from a well-performing model provides a blueprint for task-specific model compression. By identifying which feature relationships are essential for a particular task, IGBO can guide the design of smaller, specialized models that retain performance while eliminating unnecessary complexity. This represents a shift from post-hoc explanation toward proactive model design informed by interpretability constraints.

\section{Conclusion and Future Work}
We have introduced IGBO, a novel framework for training machine learning models to adhere to domain-specific interpretability constraints. By formalizing these constraints via a DAG acting on a differentiable Relative Importance Score $H_k(\mathbf{X}, \theta)$, and by deriving a geometrically motivated update rule that dynamically decomposes task and interpretability gradients, IGBO provides a principled method to navigate the accuracy-interpretability trade-off. Theoretical analysis proves convergence to Pareto-stationary points via projection mapping $\mathcal{P}$ (Theorems \ref{thm:aligned_complete}--\ref{thm:convergence_pareto}) and establishes robustness to gradient noise (Theorem \ref{thm:noise_characterization}). Our CLT-based graph construction provides statistical guarantees on DAG acyclicity and transitivity, with unconditional guarantees for the median threshold and conditional guarantees for higher confidence levels.

\textbf{Limitations and Future Work:}
\begin{itemize}
    \item \textbf{Computational Overhead:} The computation of $H_k(\mathbf{X}, \theta)$, involving multiple gradient evaluations for the path integral, adds overhead per training step. While parallelizable, future work should investigate more efficient, approximate attribution methods that retain differentiability but are cheaper to compute.
    
    \item \textbf{Optimal Path Oracle:} We have described the architecture and training procedure for an Optimal Path Oracle to mitigate the OOD problem in TIG computation. We leave the full implementation, training, and empirical evaluation of the Oracle to future work. Integrating the Oracle into the IGBO training loop is a promising direction for improving attribution robustness.
    
    \item \textbf{DAG Specification and Human Expertise:} The current framework requires the DAG and intervals $[\epsilon, \delta]$ to be specified by an expert. In scenarios where gradient access is limited or models are treated as black-box computational functions, human expert annotations can directly provide feature importance rankings. Future work could explore using pairwise comparison data from domain experts to construct the DAG, or leveraging a teacher model's importance statistics to automate the process.
    
    \item \textbf{Task-Specialized Model Compression:} Beyond interpretability itself, IGBO enables a pathway toward task-specific model optimization. Given a large model that performs well on a particular task, the structured interpretability DAG extracted under IGBO captures task-relevant feature dependencies. This structure can guide the training of a smaller, task-specialized model that preserves performance while discarding unnecessary capacity. Unlike parameter-efficient fine-tuning approaches that adapt large models in-place, this perspective emphasizes constructing compact models optimized explicitly for a single task, using interpretability as a mechanism for structured knowledge transfer.
    
    \item \textbf{Theoretical Refinements:} Further analysis could establish non-asymptotic convergence rates or study the geometry of the Pareto frontier discovered by IGBO.
\end{itemize}

In conclusion, IGBO offers a rigorous and practical pathway to move beyond post-hoc explanations, enabling the creation of models whose internal decision-making processes are inherently guided by, and thus more aligned with, human-understandable semantic structures.

\bibliographystyle{IEEEtran}
\bibliography{refs}

\end{document}